\documentclass[conference, 10pt, letterpaper]{IEEEtran}
\IEEEoverridecommandlockouts
\usepackage{cite}
\usepackage{amsmath,amssymb,amsfonts}
\usepackage{algorithmic}
\usepackage{graphicx}
\usepackage{tabularx}
\usepackage{textcomp}
\usepackage{xcolor}
\usepackage{colortbl}
\usepackage{romannum}
\def\BibTeX{{\rm B\kern-.05em{\sc i\kern-.025em b}\kern-.08em
    T\kern-.1667em\lower.7ex\hbox{E}\kern-.125emX}}
\begin{document}
\renewcommand{\tabularxcolumn}[1]{m{#1}}
\title{Language Augmentation in CLIP for Improved Anatomy Detection on Multi-modal Medical Images
% Augmented Language CLIP for Improved Anatomy Detection on Multi-modal Medical Imaging Data

}

\author{
    \IEEEauthorblockN{Mansi Kakkar\IEEEauthorrefmark{1}, Dattesh Shanbhag\IEEEauthorrefmark{2}, Chandan Aladahalli\IEEEauthorrefmark{2} and Gurunath Reddy M\IEEEauthorrefmark{2}}
    \IEEEauthorblockA{\IEEEauthorrefmark{1}Indian Institute of Technology Madras, \IEEEauthorrefmark{2}GE HealthCare \\ mansikakkar97@gmail.com, \{Dattesh.Shanbhag, Chandan.Aladahalli, GurunathReddy.M\}@gehealthcare.com}}

\maketitle
%%%%%%%%%%%%%%%%%%%%%%%%%%%%
\begin{abstract}
Vision-language models have emerged as a powerful tool for previously challenging multi-modal classification problem in the medical domain. This development has led to the exploration of automated image description generation for multi-modal clinical scans, particularly for radiology report generation. Existing research has focused on clinical descriptions for specific modalities or body regions, leaving a gap for a model providing entire-body multi-modal descriptions. In this paper, we address this gap by automating the generation of standardized body station(s) and list of organ(s) across the whole body in multi-modal MR and CT radiological images. Leveraging the versatility of the Contrastive Language-Image Pre-training (CLIP), we refine and augment the existing approach through multiple experiments, including baseline model fine-tuning, adding station(s) as a superset for better correlation between organs, along with image and language augmentations. Our proposed approach demonstrates $\mathbf{47.6\%}$ performance improvement over baseline PubMedCLIP.
\end{abstract}
%%%%%%%%%%%%%%%%%%%%%%%%%%%%%%%%
% \begin{IEEEkeywords}
% vision-language models, CLIP, multi-modality, multi-label classification, prompt engineering.
% \end{IEEEkeywords}
%%%%%%%%%%%%%%%%%%%%%%%%%%%%%%%%%
%%Inserting Block Diagram
\begin{figure*}[htbp]
    \centering
    \includegraphics[width = \textwidth]{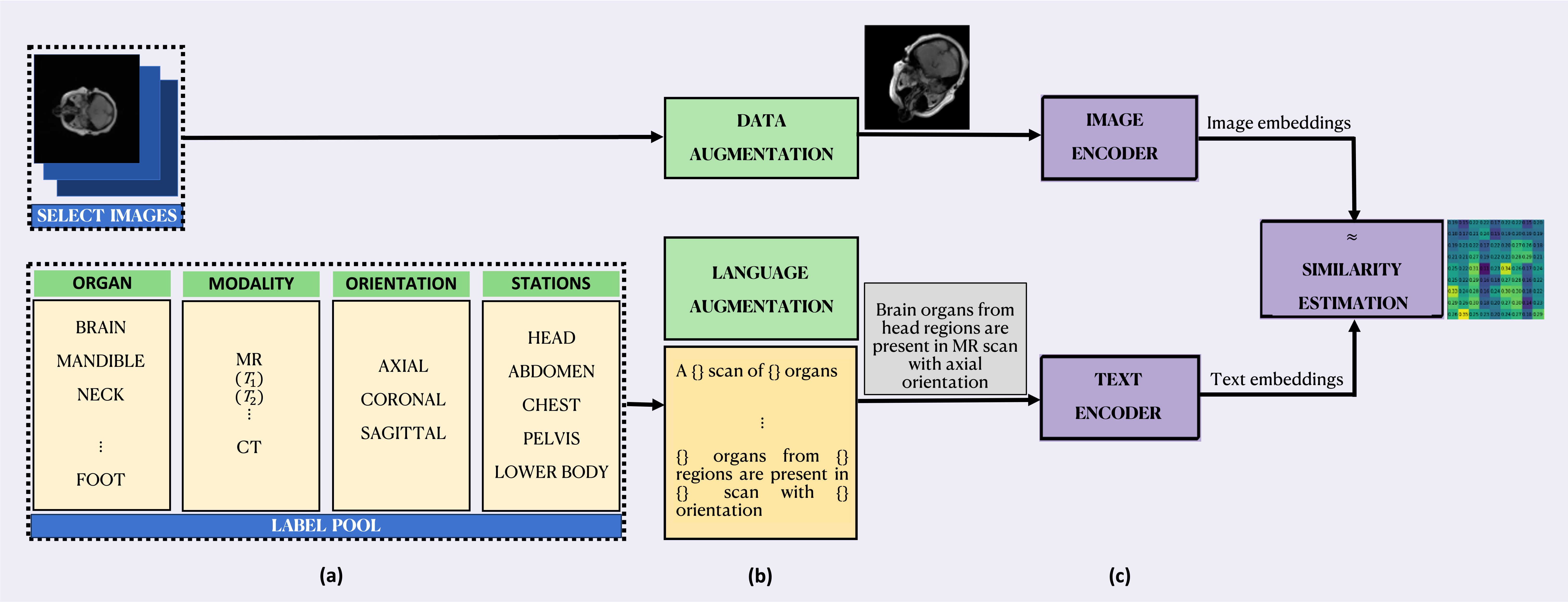}
    \caption{Pipeline of our approach for anatomy classification. (a) Dataset creation -- multi-modal anatomy dataset creation with label pools for detailed caption of images, (b) pre-processing -- image augmentation and language augmentation, involving set of labels from label pool passing through a manual prompt phrasing system providing $10$ different sets of prompts for each image, and (c) baseline model -- PubMedCLIP model with ViT-B/32 as vision encoder and text tokenizer as text encoder will be fine-tuned over these images to give us our proposed model for anatomy detection.}
    \label{Figure:1}
\end{figure*}
%%%%%%%%%%%%%%%%%%%%%%%%%%%%%
\section{Introduction}
Over the last decade, deep learning models from CNNs to Vision Transformers (ViT) \cite{b15} have gained prominence in aiding clinicians in performing their medical imaging studies efficiently through consistent image acquisition, reconstruction and AI-assisted reporting. Recent advances in Large Language Models (LLMs) have led to the integration of vision and text encoders, giving rise to Vision-language Models (VLMs), incorporating semantic descriptions into medical image analysis, and correlating textual information with image features. State-of-the-art models such as CLIP \cite{b1}, ALIGN \cite{b2}, BASIC \cite{b3}, and LiT \cite{b4} have shown remarkable results in cross-modal search using zero-shot classification and image-to-text and text-to-image applications over the multi-modal datasets \cite{b5,b6}. In the medical context, where images often require descriptive textual interpretation, VLMs become crucial for relating visual features to clinical findings, supporting prognosis and diagnosis by leveraging multi-modal data.

This work explores VLMs to describe organs and anatomical regions in multi-modal radiology images. This is crucial to describe the anatomical context for image acquisition and reporting. Given challenges in semantically segmenting diverse anatomies across each image, a vision encoder-only approach may not be ideal due to wide patient pose, orientation, and coverage variations. Text-based descriptors for organs and regions is much more attractive due to simplified labelling and the potential for capturing detailed hierarchical information. We utilize CLIP for its proven effectiveness in addressing multi-modal challenges, specifically within the realm of medical data \cite{b7}. For clinical applications, CLIP has been refined to PubMedCLIP using multi-modal medical image-text pairs, demonstrating organ-specific vision and language embeddings \cite{b9}. Since PubMedCLIP has been trained over images obtained from publications, it typically fails to provide good results on pristine medical image data.

In this work, we evaluated a methodology to fine-tune  PubMedCLIP model over clinical imaging data for labelling organs and anatomical regions (stations). Further, this paper demonstrates the importance of text prompts \cite{b1, b10} drawing inspiration from LLMs, where manipulating the textual modality is a common practice. We use VLMs inherent data manipulation capability to our advantage for multi-modal classification. Our key contributions include, (1) analysing PubMedCLIP model over multi-modal, multi-label classification on pristine clinical single slice datasets for anatomy description; (2) fine-tuning PubMedCLIP on approximately $4000$ clinical scans to achieve enhanced performance for multi-label anatomy detection; (3) showcasing enhanced model performance using data augmentations for both images and text phrases.
%%%%%%%%%%%%%%%%%%%%%%%%%%%%%%%%%%%%%%%%%%%%%%%%%%
\section{Methods}
The proposed multi-modal anatomy detection framework is shown in Fig. \ref{Figure:1}.
%%%%%%%%%%%%%
\subsection{Base model}\label{II-A}
In this study, we have used PubMedCLIP as our base model which is CLIP fine-tuned over ROCO (Radiology Objects in Context) dataset \cite{b12}. ROCO dataset comprises approximately $82$K radiology images and captions from diverse modalities, typically image-caption pairs from PubMed. ROCO dataset has provided promising outcomes in clinical use cases \cite{b5, b12}. Our experiments have demonstrated the ineffectiveness of PubMedCLIP describing organs, possibly due to the presence of various artefacts like figures, portraits, digital arts, and illustrations in the ROCO dataset, unrelated to clinical medical scans. Moreover, MR and CT scans in ROCO dataset have highlighted or marked artefacts, and an imbalance towards some organs. Due to these limitations, we further fine-tune PubMedCLIP with additional clean multi-modal clinical images and labels.
%%%%%%%%%%%%%%%%
\subsection{Data description}\label{II-B}
We fine-tuned the base model for our study using data from various sources, including in-house and clinical open-source datasets (TotalSegmentator \cite{b11}). The data includes modalities (MR and CT), orientations (axial, coronal, and sagittal), and covers different organs. MR includes data across protocols: T1, T2, FLAIR, DWI, ADC and STIR. The images were taken from five different stations of the human body: head, chest, abdomen, pelvis, and lower body. The image labels have been distributed over $20$ organ labels--brain, mandible, neck, shoulder, humerus, elbow, forearm, wrist, hand, lungs, heart, liver, kidneys, intestine, pelvic bone, thigh, knee, leg, ankle, and foot. We included additional labels, such as modality (including protocols), orientation, and stations/regions along with the organs, to generate text captions similar to the ROCO dataset. The overall training dataset contains single-slice images, captions, and organ labels. The total training dataset size was $4994$ images, comprising $3995$ for model training and $999$ validation images. We used two datasets to test the approach: Set \#1 comprising $262$ MR and CT in-house images and Set \#2 comprising of open-source visible human project \cite{b13} with $650$ CT images.
%%%%%%%%%%%%%%
\subsection{Augmentation and pre-processing of images}\label{II-C}
To mimic the data variety encountered in clinical practices, we include data from different modalities and apply augmentations. We have used these augmentations: 1) histogram manipulations (CLAHE, contrast enhancement using PIL library \cite{b14} and gamma correction), 2) rotation $(-180 \text{ to } 180^\circ)$ and 3) translation $(-100 \text{ to } 100 \text{ pixels})$. Using combinations of these augmentations, we generate $10$ different images for each image. CLIP model itself performs some data augmentations involving random resizing, random noise, gaussian blur, colour jittering, horizontal and vertical flips and arbitrary rotation. Our augmentations yielded better results than baseline methods, further discussed in the results section.
%%%%%%%%%%%%%%%
\subsection{Language augmentation for training}\label{II-D}
Augmentation on text prompt phrases and fine-tuning make the model task-specific, generalized for different modalities, and more robust to distribution shift. A complete prompt will describe modality, orientation, station and anatomy. Complete prompt diversity is obtained by shuffling the entities. To enhance model's robustness, we augmented text prompts with incomplete information, such as missing station or organ details. See Table \ref{Table 1.} for examples.
%%%%%%%%%%%%%%%%
\begin{table}[b]
\caption{Examples of prompt texts given as captions for images for training and validation datasets}
\begin{center}
\begin{tabularx}{\linewidth}{|>{\hsize=0.1\hsize}X|>{\hsize=1.9\hsize}X|}
\hline
\multicolumn{2}{|c|}{\cellcolor[HTML]{FEB125}\textbf{Prompts}}\\
\hline
\cellcolor[HTML]{FFE1A6}1. & \cellcolor[HTML]{FFE1A6}A \textit{\{orientation\}} oriented \textit{\{modality\}} image of \textit{\{organ\}} organs belong to \textit{\{station\}} region\\
& \textbf{E.g.,} A sagittal oriented MR T2 image of knee organs belong to lower body region\\
\hline
\cellcolor[HTML]{FFE1A6}2. & \cellcolor[HTML]{FFE1A6}An image of \textit{\{orientation\}} \textit{\{modality\}} scan consisting of \textit{\{organ\}} organs\\
& \textbf{E.g.,} An image of axial CT scan consisting of liver, intestine organs\\
\hline
\end{tabularx}
\label{Table 1.}
\end{center}
\end{table}
%%%%%%%%%%%%%%%
\subsection{Fine tuning}\label{II-E}
We experimented with fine-tuning PubMedCLIP, with encoders from two architectures: ResNet50 \cite{b16} and ViT–B/32 \cite{b15}. We proceeded further with ViT–B/32 since it provided comparatively better performance. We retained the baseline text tokenizer, and trimmed longer captions while zero padding the shorter ones, according to CLIP's maximum acceptable text length of $76$. Moreover, we used the baseline loss function,
\begin{equation}    \label{eq:1}
    L = \lambda H(\hat{y}_{vision} , Y)+(1-\lambda)H(\hat{y}_{text} , Y),
\end{equation}
%%%%%%%%
where $H$ is the cross entropy loss, $Y$ is the set of labels, and $\lambda = 0.5$ for equal weightage to vision and text losses. Further, we explored multiple training approaches including expanding the input training data. The resulting performance is discussed in the experiments section.
%%%%%%%%%
\subsection{Zero--shot prediction}\label{II-F}
After fine-tuning the model on our multi-modal anatomy dataset, we set up a zero-shot classifier, independently for $5$ station labels and $20$ organ labels. Logits are computed between fixed dimensional text and image feature vectors, obtained by sending text prompt to the text encoder and image to the image encoder, respectively, as shown in Fig. \ref{Figure:1}. Zero-shot classification is accomplished by using
%%%%%%%%%%
\begin{equation}    \label{eq:2}
    logits_m = [I_m\cdot T_1, \dots , I_m\cdot T_k, \dots , I_m\cdot T_{20}],
\end{equation}
%%%%%%%%%%
\begin{equation}    \label{eq:3}
    \hat{y}^1_m = \operatorname*{arg\,max} \; \text{softmax}(logits_m),
\end{equation}
%%%%%%%%%%%
to get the top prediction, where $I_m$ is individual image embedding, $T_1,\dots,T_{20}$ are text embedding for 20 labels prompts, and $\hat{y}_m^1$ is the top prediction. A similar computation is done for stations. To ensure computational accuracy, we verify if the predicted label exists within the set of target labels by
\begin{equation} \label{eq:4}
    \text{accuracy} = \frac{\sum_{i=1}^N\text{\# correct matches for image } i}{N} ,
\end{equation}
where $N$ is the size of the total test dataset.
%%%%%%%%%%%%%%%%%%%%%%%%%%%%%%%%%%%%%%%%%%%%%%%%%%
\section{Experiments}
\begin{table}[t]
\caption{List of experiments performed}
\begin{center}
\begin{tabularx}{\linewidth}{|>{\hsize=0.9\hsize\arraybackslash}X|>{\hsize=0.1\hsize\arraybackslash}c|}
\hline
\cellcolor[HTML]{FFCE93}\textbf{Experiments}&\cellcolor[HTML]{FFCE93}\textbf{Augmentations}\\
\hline
\cellcolor[HTML]{FFFFFF}\textbf{Exp 1. (PMC) }Tested baseline model (PubMedCLIP with ViT-B/32 vision encoder).& \cellcolor[HTML]{FFFFFF}No\\
\hline
\cellcolor[HTML]{FFFFFF}\textbf{Exp 2. (PMC-M) }Fine-tuned baseline model over multi-modal anatomy dataset of 3995 training images and corresponding captions involving organ(s), modality, and orientation labels (see Section \ref{II-B}).&\cellcolor[HTML]{FFFFFF}No\\
\hline
\cellcolor[HTML]{FFFFFF}\textbf{Exp 3. (PMC-MS) } Revised captions for better region segregation by including station labels. For eg., labels like brain and knee do not get paired in an axial scan.&\cellcolor[HTML]{FFFFFF}No\\
\hline
\cellcolor[HTML]{FFFFFF}\textbf{Exp 4. (PMC-MSA) }Performed augmentations over training images and  captions as described in Section \ref{II-C}, and Section \ref{II-D}, respectively. &\cellcolor[HTML]{FFFFFF}Yes\\
\hline
\end{tabularx}
\end{center}
\label{Table:2}
\end{table}
%%%%%%%%%%%%%%
\begin{figure*}[t]
    \centering
    \includegraphics[width=\textwidth]{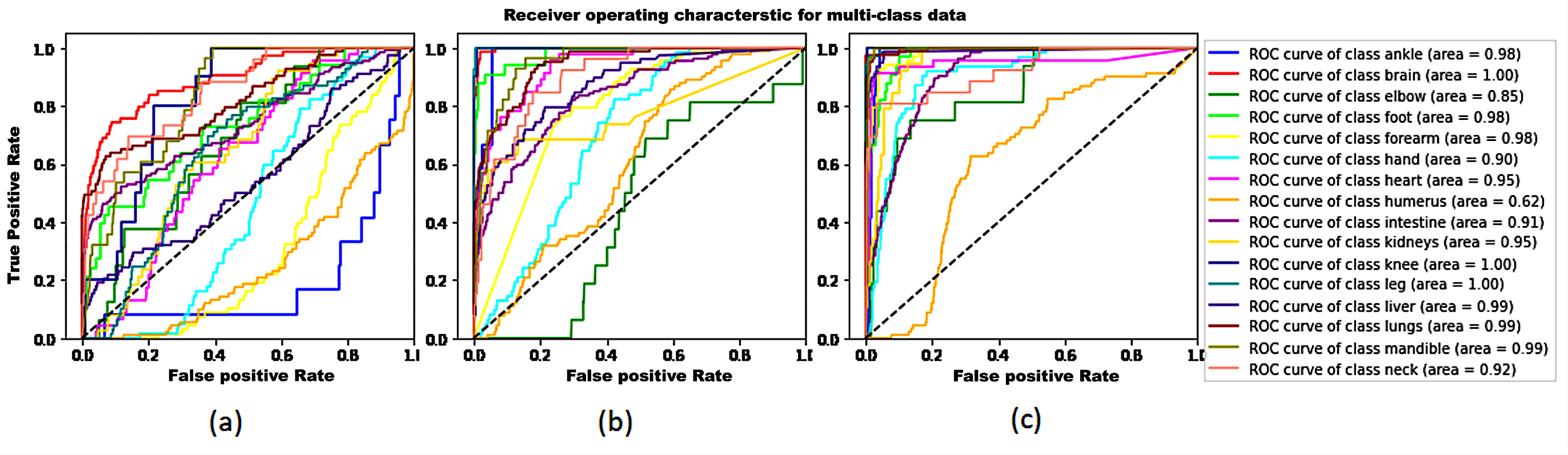}
    \caption{AUC-ROC curves for test dataset (visible human project) across different models: (a) result for PMC (baseline), (b) result for PMC-M (fine-tuning over mutli-modal anatomy dataset), and (c) result for PMC-MSA (model with text and image data augmentations). The AUC values are for proposed approach. We receive good AUC values for all organs except humerus (AUC = $0.62$), probably due to imbalance towards other limbs as compared to humerus}
    \label{Figure:3}
\end{figure*}
%%%%%%%%%%%%%%%
\begin{figure}[t]
    \centering
    \includegraphics[width=\columnwidth]{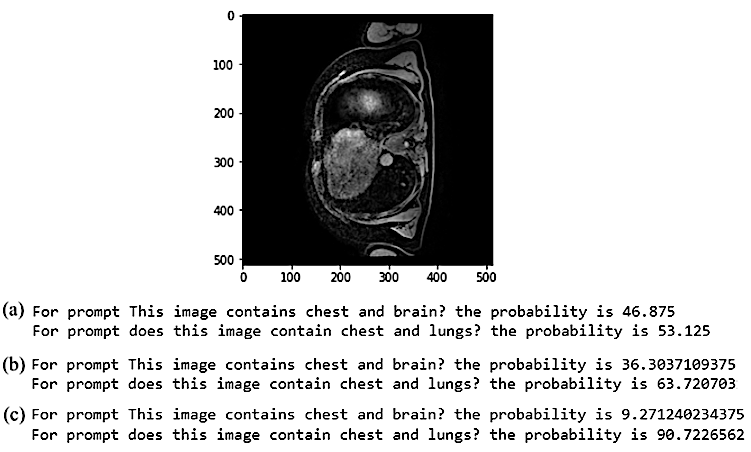}
    \caption{Examples of performance for different models: (a) result for PMC (baseline), (b) result for PMC-MS, and (c) result for PMC-MSA (proposed approach). Showcasing our approach outperforming the baseline}
    \label{Figure:2}
\end{figure}
%%%%%%%%%%%%%%%%
\begin{table}[b]
    \centering
    \caption{Zero-shot accuracy comparison among all experiments. Set $\#1$--in-house MR and CT test dataset and Set $\#2$--open source CT dataset}
    \begin{tabularx}{\columnwidth}{|m{0.7cm}|m{0.97cm}|m{1.2cm}|m{1.0cm}|m{1.0cm}|m{1.36cm}|}
     \hline
     \cellcolor[HTML]{C0E470}\textbf{Label Type}&\cellcolor[HTML]{C0E470}\textbf{Dataset}&\cellcolor[HTML]{C0E470}\textbf{PMC [Baseline]}&\cellcolor[HTML]{C0E470}\textbf{PMC-M}&\cellcolor[HTML]{C0E470}\textbf{PMC-MS}&\cellcolor[HTML]{C0E470}\textbf{PMC-MSA [proposed approach]}  \\
     \hline
     \cellcolor[HTML]{C0E470}&\cellcolor[HTML]{E8FABD}Set \#1 (MR)&\centering\arraybackslash\cellcolor[HTML]{E8FABD}54\%&\centering\arraybackslash\cellcolor[HTML]{E8FABD}62\%&\centering\arraybackslash\cellcolor[HTML]{E8FABD}57\%&\centering\arraybackslash\cellcolor[HTML]{E8FABD}\textbf{91.62\%}\\
     \cellcolor[HTML]{C0E470}\textbf{Organ labels}&Set \#1 (CT)&\centering\arraybackslash33\%&\centering\arraybackslash70.8\%&\centering\arraybackslash63\%&\centering\arraybackslash\textbf{81.25\%}\\
     \cellcolor[HTML]{C0E470}&\cellcolor[HTML]{E8FABD}Set \#2 (CT)&\centering\arraybackslash\cellcolor[HTML]{E8FABD}24\%&\centering\arraybackslash\cellcolor[HTML]{E8FABD}58.5\%&\centering\arraybackslash\cellcolor[HTML]{E8FABD}47\%&\centering\arraybackslash\cellcolor[HTML]{E8FABD}\textbf{81\%}\\
     \hline
     \cellcolor[HTML]{C0E470}\textbf{Station labels}&Set \#1 (MR+CT)&\centering\arraybackslash52\%&\centering\arraybackslash49\%&\centering\arraybackslash65\%&\centering\arraybackslash\textbf{75\%}\\
     \cellcolor[HTML]{C0E470}&\cellcolor[HTML]{E8FABD}Set \#2 (CT)&\centering\arraybackslash\cellcolor[HTML]{E8FABD}\cellcolor[HTML]{E8FABD}45\%&\centering\arraybackslash\cellcolor[HTML]{E8FABD}41\%&\centering\arraybackslash\cellcolor[HTML]{E8FABD}60\%&\centering\arraybackslash\cellcolor[HTML]{E8FABD}\textbf{76\%}\\
     \hline
    \end{tabularx}
    \label{Table:3}
\end{table}
%%%%%%%%%%%%%%
\subsection{Annotation and Implementation}
We annotated the datasets in-house, consisting of labels for organs, modality, orientation and stations. The annotations were reviewed with a trained radiologist and a clinician. The training was performed on NVIDIA DGX A100 Tensor Core GPU. We conducted four experiments using different configurations: (1) PubMedCLIP (PMC), (2) PMC fine-tuned over a multi-modal anatomy dataset (PMC-M), (3) PMC-M with additional stations (PMC-MS), and (4) PMC-MS with image and text augmentations (PMC-MSA). The details are provided in Table \ref{Table:2}.
\subsection{Evaluation}
All experiments are assessed through zero-shot classification, analyzing the top accuracy. Further, we visualize the performance of models PMC, PMC-M, and PMC-MSA, through a One-vs-the-Rest (OvR) AUC-ROC curve using \textit{Scikit-Learn}.
%%%%%%%%%%%%%%%%%%%%%%%%%%%%%%%%%%%%%%%%%%%%%%%%%
\section{Results and Analysis}
The results for all experiments are given in Table \ref{Table:3}. Our proposed approach gives an overall average enhancement of $47.6\%$ for organ detection, and $27\%$ for station detection in comparison to the baseline model. After fine-tuning the baseline model over multi-modal anatomy dataset, the organ prediction performance is improved in PMC-M, but gave a poor performance for station prediction due to its reliance on the ROCO dataset for station information. In PMC-MS, providing detailed station information improved station prediction, but organ prediction accuracy dropped, primarily attributed to confusion between station and organ (no strong correlation observed). Further, in PMC-MSA, this confusion is reduced by data augmentation and text prompt diversity, enhancing both organ and station prediction performance. Fig. \ref{Figure:3} shows the OvR AUC-ROC curve for Set \#2 for different organs across the models. Our proposed model performs well for all organs, but due to data imbalance, it gives comparatively low performance for humerus. Furthermore, a comparison among PMC, PMC-MS, and PMC-MSA is performed and the result is shown in Fig. \ref{Figure:2}. Given two sets of prompts--one with correct organ-station correlation and the other with a contradiction, our approach outperforms the baseline model. Notably, for the correct and the false prompts--our approach predicts $90.7\%$ and $9.3\%$ scores, respectively, compared to the baseline scores of $53.1\%$ and $46.9\%$.
%%%%%%%%%%%%%%%%%%%%%%%%%%%%%%%%%%%%%%%%%%%%%%%%%%
\section{Conclusion}
Our research demonstrates the effectiveness of utilizing station data to establish correlations between organs, resulting in a notable improvement in accuracy. Along with it, the integration of image and text prompt augmentations significantly improves model performance. The integration of CLIP, coupled with strategic data augmentations, has notably enhanced accuracy and reliability in multi-modal organ-related studies, making a step forward in the realm of medical imaging.

We have made efforts to address the data imbalance in the ROCO dataset. Still, there is room for improvement, particularly when assessing the limbs' performance compared to other organs. As a future step, we could learn class-specific text tokens to eliminate manual text augmentations to further improve the model accuracy. 
%%%%%%%%%%%%%%%%%%%%%%%%%%%%%%%%%%%%%%%%%%%%%%%%%%%
\section*{ACKNOWLEDGMENT}
All authors would like to sincerely thank clinical experts Dr. Patil and Dr. Pakhi Sharma for assisting with the annotation of the multi-modal anatomy dataset. We also thank Dr. Deepa Anand for her assistance in data labelling. 
%Their invaluable insights have significantly enriched the quality of our work.

\end{document}